\documentclass[10pt,twocolumn,letterpaper]{article}

\usepackage[pagenumbers]{iccv} 

\definecolor{iccvblue}{rgb}{0.21,0.49,0.74}
\usepackage[pagebackref,breaklinks,colorlinks,allcolors=iccvblue]{hyperref}

\title{EdgeDoc: Hybrid CNN-Transformer Model for Accurate Forgery Detection and Localization in ID Documents}

\author{Anjith George and  Sébastien Marcel \\
Idiap Research Institute, Switzerland\\
{\tt\small \{anjith.george,sebastien.marcel\}@idiap.ch}\\}

\begin{document}
\maketitle
\begin{abstract}
The widespread availability of tools for manipulating images and documents has made it increasingly easy to forge digital documents, posing a serious threat to Know Your Customer (KYC) processes and remote onboarding systems. Detecting such forgeries is essential to preserving the integrity and security of these services. In this work, we present EdgeDoc, a novel approach for the detection and localization of document forgeries. Our architecture combines a lightweight convolutional transformer with auxiliary noiseprint features extracted from the images, enhancing its ability to detect subtle manipulations. EdgeDoc achieved third place in the ICCV 2025 DeepID Challenge, demonstrating its competitiveness. Experimental results on the FantasyID dataset show that our method outperforms baseline approaches, highlighting its effectiveness in real-world scenarios. Project page : \url{https://www.idiap.ch/paper/edgedoc/}
\end{abstract}    
\section{Introduction}
\label{sec:intro}

The widespread adoption of digital KYC processes in financial services has introduced new security risks, as forged identity documents can be injected or physically replayed to bypass verification systems. Advances in image generation \cite{borji2022generated} and editing have made such forgeries increasingly realistic and harder to detect, especially when involving subtle text manipulations. Although recent methods \cite{wang2025forensics,zhao2021deep,munoz2025exploring,wang2025research,sanchez2024few,tapia2024first} have improved manipulation detection, many overlook fine-grained document-level forgeries.

A major challenge in the domain of forgery detection is the lack of model generalization. Identity documents exhibit substantial variation in design across different regions, making it difficult to develop a single model capable of effectively generalizing across all layout types. Moreover, the nature of forgery attacks can differ significantly: some may involve alterations to textual content, others may target facial images, and some may modify both. Detecting forgeries becomes particularly challenging when the tampered region is relatively small. An additional limitation is the requirement for large-scale datasets to train robust models. However, the sensitive and personally identifiable nature of identity documents poses significant constraints on data collection and sharing, limiting the availability of comprehensive and realistic training datasets.

Motivated by the significance of the problem and its inherent challenges, we introduce EdgeDoc, a novel method for document manipulation detection. The proposed model employs a lightweight hybrid architecture that integrates convolutional and transformer-based components, enabling simultaneous classification and forgery localization. Designed to perform effectively with a limited number of training samples, EdgeDoc achieves competitive results on a public benchmark challenge.

\section{Proposed Method}
The proposed method, EdgeDoc, is a hybrid model that combines the strengths of the TruFor framework \cite{guillaro2023trufor} with a custom lightweight architecture inspired by EdgeFace \cite{george2024edgeface}. Given the limited availability of training data, training a model from scratch is impractical. To address this, we leverage the NoisePrint representation extracted via the TruFor pipeline, which serves as a source of localized anomaly cues. This NoisePrint is fused with the original image to enable patch-wise interaction within a convolutional-transformer architecture, facilitating both manipulation detection and localization. The details of the proposed approach are presented in the following subsections.

\begin{figure*}[ht] 
               \centering
               \includegraphics[width=\textwidth]{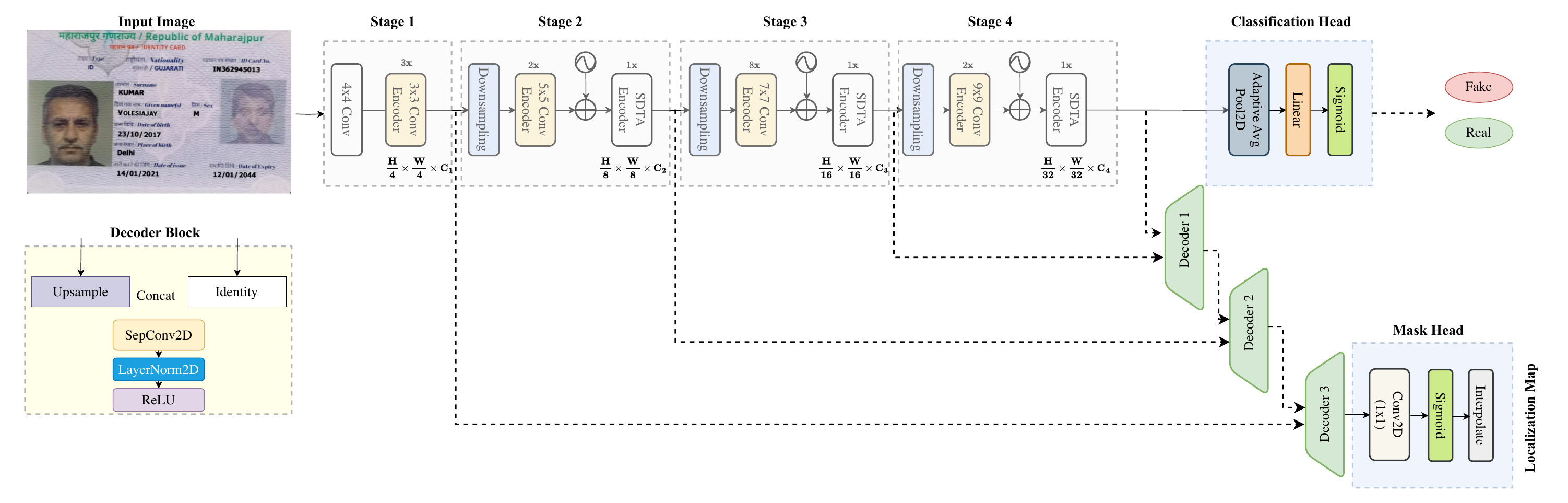}
               \caption{Architecture of the proposed EdgeDoc framework}
               \label{fig:image_architecture} 
\end{figure*}

\subsection{Device fingerprint extraction}

In \cite{cozzolino2019noiseprint}, the authors introduced NoisePrint, a neural network designed to extract a camera model fingerprint. The model is trained to suppress scene-related artifacts while enhancing camera model-specific patterns. Training is performed using a Siamese architecture, where image patch pairs from the same camera are treated as positive examples, and those from different cameras as negative examples. Building on this, the TruFor framework \cite{guillaro2023trufor} improved the method by incorporating a transformer-based fusion module that combines RGB information with an enhanced version of NoisePrint (called NoisePrint++). Forgery detection is approached as the identification of deviations from the expected regularity in the image. The framework outputs an integrity score, a localization map, and a confidence score, helping to identify potentially error-prone regions with higher precision.

\subsection{Convolutional-Transformer Hybrid Network}

EdgeFace~\cite{george2024edgeface} demonstrated the effectiveness of convolutional-transformer hybrid architectures for face recognition tasks, building on the EdgeNeXt framework~\cite{maaz2022edgenext}. These architectures combine the local inductive biases of convolutional layers with the global modeling capabilities of transformers, all within a compact and computationally efficient design. Motivated by their balance of performance and efficiency, we adopt a similar lightweight hybrid architecture in the development of our model.

\subsection{EdgeDoc Architecture}

Our proposed architecture, EdgeDoc, is based on the XXS variant of the EdgeNeXt backbone. It extracts multi-scale feature maps from various stages of the network, which are then fed into a custom decoder structured in a U-Net style. The architecture of EdgeDoc is shown in Fig. \ref{fig:image_architecture}. The decoder is composed of upsampling blocks, each consisting of depthwise separable 2D convolutions, followed by 2D Layer Normalization and ReLU activations.

For classification, we utilize a bottleneck head comprising global average pooling and fully connected layers. The final segmentation mask is generated via a pointwise (1×1) convolution applied to the decoder output.

\subsection{Training Details}

The input to the network comprises two channels: the green channel of the ID image and the NoisePrint feature map.
For the classification task, binary cross-entropy (BCE) loss was employed. Localization was optimized using a composite loss function combining BCE and Dice loss \cite{milletari2016v}. A weighting factor of $\lambda = 3.0$ was applied to the mask loss component during training. The total loss function ($\mathcal{L}_{\text{total}}$) is defined as follows:

\begin{align}
\mathcal{L}_{\text{cls}} &= \text{BCE}(y_{\text{cls}}, \hat{y}_{\text{cls}}) \\
\mathcal{L}_{\text{mask}} &= \text{BCE}(y_{\text{mask}}, \hat{y}_{\text{mask}}) + \text{Dice}(y_{\text{mask}}, \hat{y}_{\text{mask}}) \\
\mathcal{L}_{\text{total}} &= \mathcal{L}_{\text{cls}} + \lambda \cdot \mathcal{L}_{\text{mask}}
\end{align}
where $\hat{y}_{\text{cls}}$ is the output from the classification head and 
$\hat{y}_{\text{mask}}$ the output from the mask head, 
$y_{\text{cls}}$ and $y_{\text{mask}}$ denote corresponding ground truths.

The model was trained using the AdamW \cite{loshchilov2017decoupled} optimizer with a weight decay of $5 \times 10^{-4}$ and a batch size of 1. The initial learning rate was set to $3 \times 10^{-4}$ and decayed according to a cosine annealing schedule over 20 epochs. The model achieving the lowest validation loss during training was selected for final evaluation. All experiments were conducted on an NVIDIA RTX 3090 GPU.

\section{Experiments}

This section presents the experimental setup, baseline methods, and the corresponding results.

\begin{figure*}[] 
               \centering
               \includegraphics[width=\textwidth]{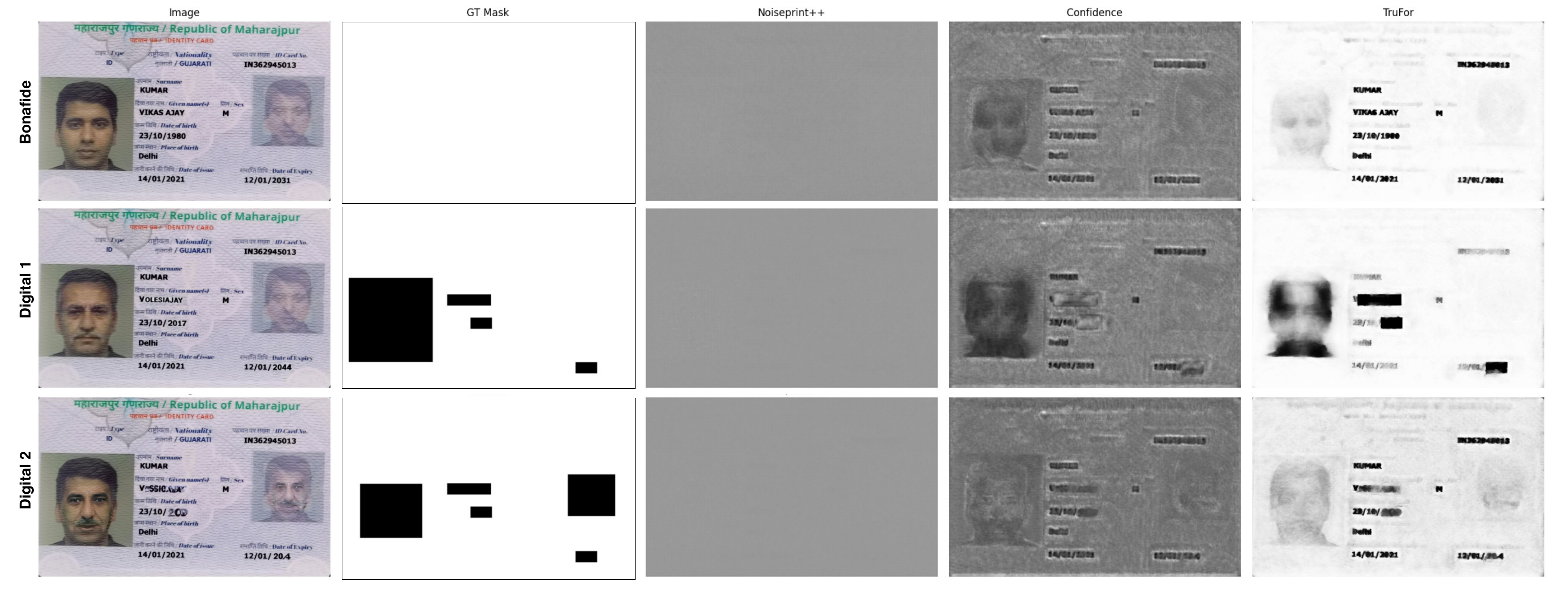}
               \caption{Sample and Ground Truth from Fantasy ID dataset, together with the NoisePrint++, Confidence and Localization results from TruFor}
               \label{fig:image_fantasy} 
\end{figure*}

\begin{figure}[] 
               \centering
               \includegraphics[width=\columnwidth]{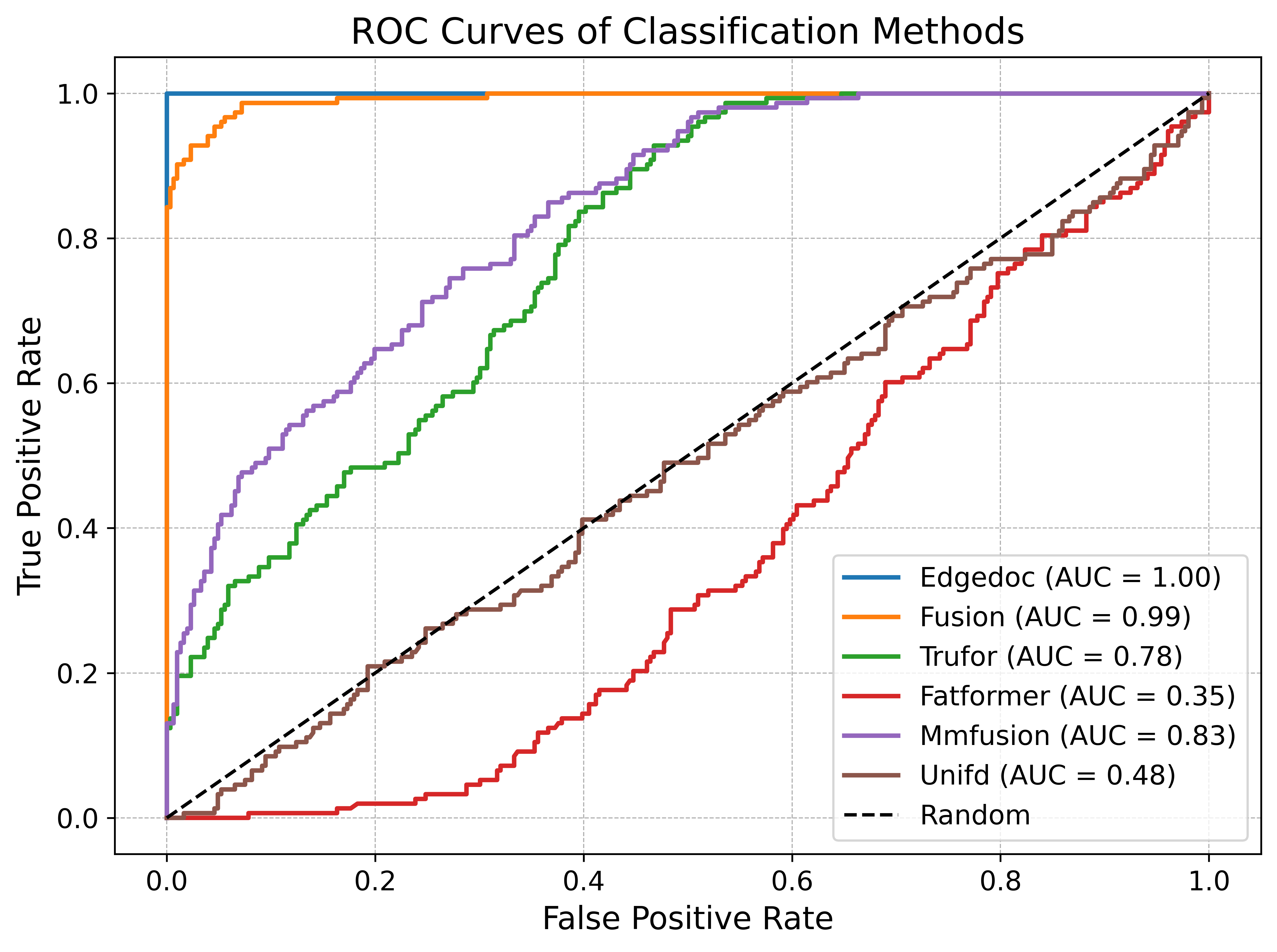}
               \caption{ROC curves for various methods from the literature on the public validation set of FantasyID dataset, along with our proposed EdgeDoc and Fusion approaches.}
               \label{fig:rocs} 
\end{figure}

\textbf{Dataset}: FantasyID \cite{korshunov2025fantasy} is a high-quality dataset developed to support research in document forgery and presentation attack detection within biometric KYC systems. It consists of two categories: bonafide identity cards and attack samples. The bonafide subset includes 262 synthetically generated fantasy ID cards featuring randomized personal information and real facial images sourced from public datasets. These cards were printed on plastic using a commercial ID card printer and captured under realistic conditions using three different imaging devices, resulting in 786 genuine images. The attack subset includes both digital manipulations created by altering facial and textual content using state-of-the-art generative models and printed manipulations, where digitally forged cards were reprinted and re-captured to simulate more sophisticated presentation attacks. Samples for bonafide and attacks are shown in Fig. \ref{fig:image_fantasy}. This dual-type attack design closely reflects real-world adversarial scenarios, offering a comprehensive and challenging benchmark for evaluating the robustness of document forgery detection algorithms. In this work, we use only the public training and validation set for the experiments.

\textbf{Metrics}: We report a comprehensive set of performance metrics for the binary classification task. The accuracy metric captures overall correctness but can be misleading under class imbalance. To address this, we include the weighted F1-score, which balances precision and recall by class weight. ROC AUC provides a threshold-independent view of true- vs. false-positive trade-offs, while the Matthews Correlation Coefficient (MCC) offers a balanced summary measure, robust even under severe skew. Together, these metrics offer a detailed evaluation of model performance across different conditions.

\textbf{Baselines}: We utilize four state-of-the-art algorithms for binary manipulation detection in images: TruFor~\cite{guillaro2023trufor}, MMFusion~\cite{triaridis2024exploring}, UniFD~\cite{ojha2023towards}, and FatFormer~\cite{liu2024forgery}. We utilize the pretrained models released by the respective authors for our experiments.

\begin{table*}[htb]
    \centering
    \caption{Performance on the public validation set of Fantasy ID}
    \label{tab:fantasy_results_table}
    \resizebox{0.65\textwidth}{!}{
\begin{tabular}{lcccccc}
\toprule
\textbf{Model} & \textbf{Accuracy} & \textbf{F1 (weighted)} & \textbf{Precision} & \textbf{Recall} & \textbf{ROC AUC} & \textbf{MCC} \\
\midrule

Fatformer & 0.34 & 0.21 & 0.33 & 0.93 & 0.35 & -0.06 \\
Mmfusion & 0.69 & 0.59 & 1.00 & 0.08 & 0.83 & 0.23 \\
Unifd & 0.33 & 0.17 & 0.33 & 1.00 & 0.48 & 0.00 \\

TruFor \cite{guillaro2023trufor} & 0.71 & 0.70 & 0.59 & 0.44 & 0.78 & 0.32 \\ \midrule

EdgeDoc & \textbf{1.00} & \textbf{1.00} & \textbf{1.00} & \textbf{1.00} & \textbf{1.00} & \textbf{1.00} \\\midrule
Fusion(EdgeDoc, TruFor ) & 0.95 & 0.95 & 0.99 & 0.87 & 0.99 & 0.90 \\
\bottomrule
\end{tabular}
    }

\end{table*}

\textbf{Experimental Results}: We trained the proposed EdgeDoc model using the training set of the Fantasy ID dataset and evaluated its performance on the corresponding validation set. In addition, we assessed several off-the-shelf baseline methods, including TruFor, for comparative analysis. The results of these evaluations are summarized in Table~\ref{tab:fantasy_results_table}, where EdgeDoc demonstrates superior performance compared to all other methods. Receiver Operating Characteristic (ROC) curves for the baselines are presented in Figure~\ref{fig:rocs}. Furthermore, we explored a fusion of EdgeDoc and TruFor using a weighted combination, which also yielded competitive results.

\textbf{ICCV 2025 DeepID Challenge Submission}
The ICCV 2025 DeepID Challenge \cite{korshunov2025deepid} represents the first competition focused on detecting synthetic manipulations, specifically injection attacks as opposed to traditional presentation attacks in identity documents. As part of the challenge, the organizers released the train and development partitions of the Fantasy ID dataset along with corresponding ground-truth labels.

Given the limited availability of publicly accessible training data, we utilized a pretrained TruFor model to extract NoisePrint maps, which provide localized cues indicative of potential manipulations. These maps, together with the original images, were used as inputs to train our custom EdgeDoc model. EdgeDoc is designed to produce both a binary classification score and a segmentation mask, thereby enabling simultaneous detection and localization of forgeries. The model was trained solely on the public training subset of the Idiap Fantasy ID dataset.

For inference, we applied a fusion strategy combining the outputs of both EdgeDoc and TruFor, specifically their classification scores and localization masks to generate a final prediction score. The performance results, including those for the individual models and their fusion, as reported on the official competition leaderboard, are summarized in Table~\ref{tab:competition_leaderboard}. While EdgeDoc and TruFor individually exhibit limited performance on the private test set, their fusion significantly improves generalization, demonstrating strong robustness to previously unseen manipulation scenarios.

\begin{table}[htb]
    \centering
    \caption{Performance Metrics from the Competition Leaderboard on the Fantasy ID and Private Test Datasets}
    \label{tab:competition_leaderboard}
    \resizebox{0.45\textwidth}{!}{
\begin{tabular}{lccc}
\toprule
\textbf{Model} & \textbf{F1 on Fantasy} & \textbf{F1 on Private} & \textbf{Aggregate F1} \\
\midrule
EdgeDoc        & 0.43 & 0.66 & 0.59 \\
TruFor \cite{guillaro2023trufor}        & 0.81 & 0.66 & 0.71 \\ \midrule
Fusion (EdgeDoc, TruFor)         & \textbf{0.96} & \textbf{0.71} & \textbf{0.79} \\
\bottomrule
\end{tabular}
    }
\end{table}

\section{Conclusions}
In this work, we present EdgeDoc, a lightweight framework for document forgery detection that leverages both original images and their corresponding NoisePrint representations. The proposed method demonstrated superior performance compared to other models on the development set of the Fantasy ID dataset. Our submission to the ICCV 2025 DeepID Challenge secured third place in the detection track, highlighting the effectiveness and competitiveness of the approach. We believe that with access to larger and more diverse training data, the performance of EdgeDoc can be further improved. To support future research and development, we will release the source code publicly.

\section{Acknowledgments}
This research was funded by the Swiss Center for Biometrics Research and Testing.

{
    \small
    \bibliographystyle{ieeenat_fullname}
    \bibliography{main}
}

\end{document}